\documentclass[letterpaper, 10 pt, conference]{ieeeconf}  

\IEEEoverridecommandlockouts                              

\usepackage{graphics} 
\usepackage{epsfig} 
\usepackage{mathptmx} 
\usepackage{times} 
\usepackage{amsmath} 
\usepackage{amssymb}  
\usepackage[at]{easylist}
\usepackage{mathtools}
\usepackage{setspace}
\usepackage{soul}
\usepackage{color}
\usepackage[binary-units=true]{siunitx}
\usepackage{xcolor}
\usepackage{mathtools}
\usepackage{float}
\usepackage[ruled,vlined,linesnumbered,noresetcount]{algorithm2e}
\usepackage[pagebackref=true,breaklinks=true,colorlinks,bookmarks=false]{hyperref}

\title{\LARGE \bf Manipulation of unknown objects via contact configuration regulation}
\author{
Neel Doshi$^{*,1}$ \quad Orion Taylor$^{*}$ \quad Alberto Rodriguez\vspace{0.2cm}\\
Massachusetts Institute of Technology\vspace{0.2cm}\\
$^*$Authors contributed equally, names listed in alphabetical order
\vspace{-0.4cm}
\thanks{
\hspace{-0.4cm}
$^1$ Neel Doshi is supported by an appointment to the Intelligence Community Postdoctoral Research Fellowship Program at the Massachusetts Institute of Technology, administered by Oak Ridge Institute for Science and Education through an interagency agreement between the U.S. Department of Energy and the Office of the Director of National Intelligence. \newline
This work was also supported by an ARA/Sponsored research award from Amazon. 
}
}

\def\*#1{\mathbf{#1}}
\def\?#1{\mathbb{#1}}

\newcommand{\secref}[1]{Section~\ref{#1}}

\newcommand{\figref}[1]{Fig.~\ref{#1}}

\newcommand{\myparagraph}[1]{\vspace{0.05in}\noindent\textbf{#1}}

\begin{document}

\maketitle

\thispagestyle{empty}
\pagestyle{empty}

\begin{abstract}

We present an approach to robotic manipulation of unknown objects through regulation of the object's \textit{contact configuration}: the location, geometry, and mode of all contacts between the object, robot, and environment.
A contact configuration constrains the forces and motions that can be applied to the object; however, synthesizing these constraints generally requires knowledge of the object's pose and geometry.
We develop an object-agnostic approach for estimation and control that circumvents this need. 
Our framework directly estimates a set of wrench and motion constraints which it uses to regulate the contact configuration.
We use this to reactively manipulate unknown planar objects in the gravity plane. 
A video describing our work can be found on our project page: \url{http://mcube.mit.edu/research/contactConfig.html}. 

\end{abstract}


\section{Introduction}
\label{sec:intro}

\begin{figure}
    \centering
    \includegraphics[width=0.9\columnwidth]{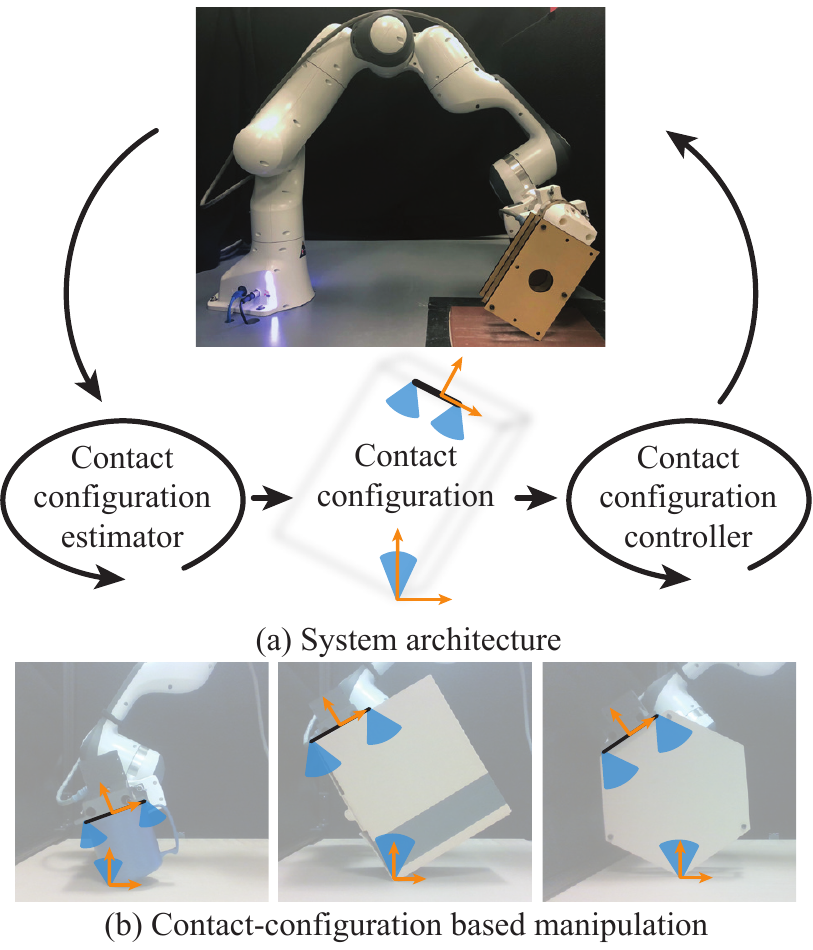}
    \vspace{-0.35cm}
    \caption{a) The robot interacts with an object by estimating and controlling its contact configuration. (b) This enables manipulation of unknown objects without explicit knowledge of their geometric and inertial properties.}
    \label{fig:intro_fig}
    \vspace{-0.5cm}
\end{figure}

Regulation of an object's \textit{contact configuration} -- the location, mode and geometry of all contacts between the object, robot, and environment -- is a fundamental abstraction of object manipulation (\figref{fig:intro_fig}).
Imagine tumbling a heavy box or tightening a screw.
Both tasks can be better described/executed by prescribing/regulating the location, geometry, and mode of all contacts. 
In these cases, the object can be sufficiently controlled without using pose, inertial, or shape information.
Even when this information is available, contact configuration regulation simplifies control, for example during  non-prehensile~\cite{chavan2020planar, hogan2020reactive} or deformable-object~\cite{she2021cable} manipulation.

As such, contact configuration regulation can be used to manipulate unknown objects (\figref{fig:intro_fig}b).
%
This is a \emph{joint estimation and control} problem.
The robot must estimate the kinematic and frictional constraints imposed by the contact configuration and regulate the contact forces and object motion accordingly. 
%
This is challenging, as not all contacts are directly observable, and the robot's control authority is limited by the underactuated mechanics of friction. 
We focus on manipulating unknown planar objects on a flat surface using robot proprioception and force/torque sensing at the wrist for feedback (\secref{sec:case_study}).
This minimal system has a diverse set of contact configurations that highlight the challenges discussed above. 

%

%
%

%

\myparagraph{Contributions} 
We manipulate the object by regulating it through a (currently predesignated) sequence of contact configurations. 
For each contact configuration, we
\begin{itemize}
    \item \textbf{Estimate} contact locations, geometries, and modes, as well as a conservative set of wrenches that can be transmitted to the object (\secref{sec:est}).
    \item \textbf{Control} the applied wrenches to maintain the contact geometries and modes, as well as the applied motions to regulate the contact locations (\secref{sec:control}).
\end{itemize}
Both the estimator and controller are object-agnostic. 
They work together to manipulate the object: the controller maintains the contact configuration to facilitate consistent estimation, and the estimated parameters improve the controller's performance (\secref{sec:joint_est_and_cont}). 
We experimentally demonstrate that our framework can reactively manipulate planar objects along a horizontal surface (\secref{sec:experiments_and_results}).

\section{Related work}
\label{sec:lit_review}

%
Prior work on contact configuration estimation and control has focused on a robot interacting with a static~\cite{her1991automated, FURUKAWA1996261} or single degree-of-freedom~\cite{karayiannidis2013model, niemeyer1997simple} unknown environment. 
%
The ability to directly observe all contacts facilitates contact configuration regulation. 
Less attention has been given to cases of robot/object/environment interactions, where
not all contacts are directly observable (e.g., during non-prehensile manipulation).
In this area, most prior research either focuses on contact configuration control assuming a known model of the world~\cite{hogan2020reactive, Hou-RSS-20, hogan2020tactile}, or contact configuration estimation assuming stable interactions~\cite{ma2021icra}.
Work on joint estimation and control either uses simplified (e.g., frictionless) models of contact \cite{Lefebvre2003polyhedral, de2007constraint}, or learns task-specific policies from data (e.g, for cable manipulation \cite{she2021cable} or part insertion \cite{dong2021icra}).
Our contribution is an object-agnostic joint estimation and control framework that reasons about all frictional interactions between the robot, object, and environment.

\myparagraph{Estimation} 
%
Several researchers have proposed methods for localizing robot/object contacts where the known geometry of the manipulator can be leveraged~\cite{bicchi1993contact, manuelli2016localizing, yu2018realtime, wang2020contact}. 
We extend these approaches to localize object/environment contacts, formalized as \textit{extrinsic} contact sensing \cite{ma2021icra}. 
Prior work on wrench constraint estimation has focused on the problem of planar pushing \cite{yoshikawa1991indentification, lynch1993estimating, zhou2018convex}. Of particular relevance is the work of Zhou et. al \cite{zhou2018convex}, who estimate the set of wrenches that can be transmitted through planar frictional contact (i.e., the limit-surface \cite{goyal1991planar}). 
We focus instead on estimating the intersection of both frictional and kinematic wrench constraints (i.e., the generalized friction cone \cite{erdmann1994representation}).

\myparagraph{Control} We highlight two  approaches that are used to simultaneously regulate force and motion: indirect force control (IFC) \cite{salisbury1980active, hogan1985impedance} and hybrid force-velocity control (HFVC) \cite{mason1981compliance, raibert1981hybrid, hou2019robust}. 
IFC regulates wrenches and motions by prescribing the interaction dynamics, while HFVC directly regulates wrenches and motions in orthogonal subspaces. 
Our controller can be viewed through both lenses. 
From the IFC perspective, it's similar to hybrid impedance control \cite{anderson1988hybrid}, where wrenches and velocities are controlled in orthogonal subspaces via impedance modulation.
From the HFVC perspective, it's similar to 
parallel force/velocity control \cite{chiaverini1993parallel}, where separate wrench and velocity controllers are summed, with priority given to the wrench controller. 
However, our approach is unique in that, instead of defining orthogonal subspaces or fixed priorities, a quadratic program determines wrench and velocity control directions and priorities online. 

\section{System overview}
\label{sec:case_study}

\begin{figure}[t]
    \centering
    \includegraphics[width=0.9\columnwidth]{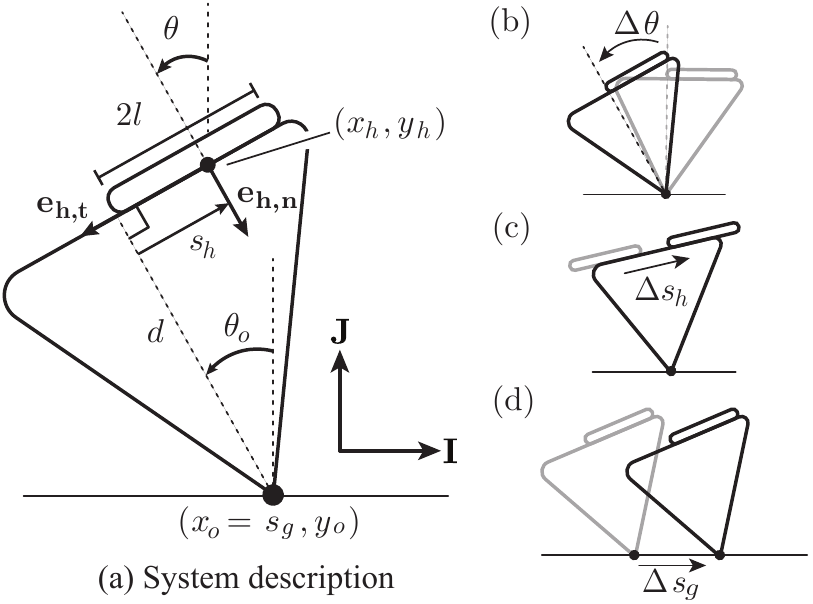}
    \vspace{-0.3cm}
    \caption{(a) System diagram with parameters, unit vectors, and state variables. (b) Pivoting while maintaining sticking at both hand and ground contacts. (c) Hand sliding while maintaining sticking contact with the ground. (d) Object sliding while maintaining sticking contact with the hand.}
    \vspace{-0.5cm}
    \label{fig:primitives}
\end{figure}

As described in \secref{sec:intro}, we consider quasi-static manipulation of unknown planar objects on a horizontal surface. 
The three main system components are: (a) the robot \textbf{hand}, which is a line of length $2l_h$, (b) the \textbf{object}, which is treated as planar convex polygon that moves in the gravity plane, and (c) the \textbf{ground}, which is a fixed horizontal line. 
The system's state consists of the planar poses of the hand $(x_h, y_h, \theta_h)$ and object $(x_o, y_o, \theta_o)$ as shown in \figref{fig:primitives}a. 
We focus on states in which the object contacts both the hand and the ground. 

Each contact configuration imposes kinematic and frictional constraints on the system. 
For example, in \figref{fig:primitives}b, sticking point contact with the ground and sticking line contact with the hand constrains the system to only admit rigid rotations of the hand and object about the ground contact (say by $\Delta \theta_o$). 
Other contact configurations (\figref{fig:primitives}c,d) can admit relative sliding between the object and the hand ($\Delta s_h$) or ground ($ \Delta s_g$).
We combine these admissible motions (i.e,  $\Delta \theta_o$, $\Delta s_h$, and $ \Delta s_g$) to explore the system's state-space. 

By the static-equilibrium condition:
\begin{align}
    \label{eq:force_static_eq}
    \sum \*{w_{net}} = \*{w_{h}} + \*{w_{g}} + \*{w_{grav}} = \*{0},
\end{align}
where $\*{w_h}$, $\*{w_g}$, and $\*{w_{grav}}$ are the wrenches exerted on the object by the hand, ground, and gravity, respectively. 
We also assume that all contact interactions follow a simple contact model with constant friction coefficients. 
Given a line segment of length $2l$ contacting a longer surface with friction coefficient $\mu$, the wrench ($f_n, f_t, \tau$) transmitted through this contact obeys: 
\begin{align}
\label{eq:friction_ineq}
-\mu f_n \leq &f_t \leq \mu f_n \quad \text{(Coulomb friction)}\\
\label{eq:torque_ineq} 
-l f_n \leq &\tau \leq l f_n \quad \ \ \text{(Line contact)}
\end{align}
Here $f_n$ and $f_t$, are the forces normal and tangential to the contact surface, and $\tau$ is net torque with respect to the center of the line. Satisfaction of the strict inequality corresponds to sticking for \eqref{eq:friction_ineq} and flush contact for \eqref{eq:torque_ineq}, while an equality admits sliding left/right for \eqref{eq:friction_ineq} and pivoting clockwise/counter-clockwise about one of the boundaries of the line segment for \eqref{eq:torque_ineq}.
 


This model allows us to describe these force-motion relationships using a series of complementarity constraints. We begin by decomposing the contact velocities, $\dot s_h$, $\dot s_g$, and the relative angle $\Delta \theta = \theta_o - \theta_h$, into their positive and negative components: 
\begin{align}
    \dot{s}_h = \dot{s}_h^+ - \dot{s}_h^-, \ \
     \Delta{\theta} = \Delta{\theta}^+ - \Delta {\theta}^-, \ \
    \dot{s}_g = \dot{s}_g^+ - \dot{s}_g^-
\end{align}
We can now relate the hand contact wrench to the relative velocity/orientation of the hand with respect to the object: 
\begin{align}
    \label{eq:slide_hand_1}
    0 \leq \mu_h f_{h,n} + f_{h,t}  \ \  &\perp \ \  \dot{s}_h^+ \geq 0 \\
    \label{eq:slide_hand_2}
    0 \leq \mu_h f_{h,n}  - f_{h,t} \ \  &\perp \ \  \dot{s}_h^- \geq 0 \\ 
    \label{eq:pivot_hand_1}
    0 \leq l_h f_{h,n}  -\tau_h \ \  &\perp \ \  \Delta \theta^+ \geq 0 \\
    \label{eq:pivot_hand_2}
     0 \leq l_h f_{h,n} + \tau_h \ \  &\perp \ \ \Delta \theta^- \geq 0
\end{align} 
The $\perp$ symbol is a shorthand, where $0 \leq a \perp b \geq 0$ implies $a \geq 0$, $b \geq 0$, and $ab = 0$. Moreover, $\*{f_h}$ and $\tau_h$ are the force and torque exerted by the hand, $\mu_h$ is the friction coefficient at the hand contact, and the subscripts $n$ and $t$ indicate the normal and tangential directions to the hand contact surface. 
Equations \eqref{eq:slide_hand_1} and \eqref{eq:slide_hand_2} relate the forces applied by the hand to relative sliding between the hand and object. Similarly, equations \eqref{eq:pivot_hand_1} and \eqref{eq:pivot_hand_2} relate the torque applied by the hand to the relative rotation between the hand and object. 

We impose similar constraints at the ground contact: 
\begin{align}
    \label{eq:slide_gnd_1}
    0 \leq \mu_g f_{g,J}-f_{g,I} \ \ &\perp \ \ \dot{s}_g^+ \geq 0 \\ 
    \label{eq:slide_gnd_2}
    0 \leq \mu_g f_{g,J} + f_{g,I} \ \ &\perp \ \ \dot{s}_g^- \geq 0,
\end{align}
where $\*{f_g}$ is the force exerted by the ground, $\mu_g$ is the friction coefficient at the ground contact, and and the subscripts $J$ and $I$ indicate the normal and tangential directions to the horizontal ground.
We use static equilibrium \eqref{eq:force_static_eq} to express \eqref{eq:slide_gnd_1} and \eqref{eq:slide_gnd_2}  in terms of the force exerted by the hand:
\begin{align}
    \label{eq:slide_gnd_3}
    0 \leq -\mu_g (f_{h,J}+f_{grav,J})+(f_{h,I}+f_{grav,I}) \ \ &\perp \ \  \dot{s}_g^+ \geq 0 \\ 
    \label{eq:slide_gnd_4}
    0 \leq -\mu_g (f_{h,J}+f_{grav,J}) - (f_{h,I}+f_{grav,I}) \ \ &\perp \ \ \dot{s}_g^- \geq 0
\end{align}

\section{Object-agnostic estimation}

\label{sec:est}
The estimator uses a time-history of the measured hand wrench ($\mathbf{w_{h,meas}}$) and pose ([$x_h, y_h, \theta_h]^T$) to estimate: 
\begin{itemize}
    \item The \textbf{generalized friction cone}: the set of all wrenches that can be applied to the object (\secref{subsec:wrench_cone_est}).
    \item The \textbf{contact mode}: whether each contact is sticking, sliding left, or sliding right (\secref{subsec:wrench_cone_est}). 
    \item The \textbf{contact geometry}: whether the hand contact is flush or pivoting (\secref{subsec:wrench_cone_est}).
    \item The \textbf{contact locations}: the relative sliding positions of the hand ($s_h$) and object ($s_g$), \secref{subsec:kinematic_est}.
\end{itemize}
These items constitute what is required to regulate the contact configuration. In addition, we also use this information to estimate and regulate $\theta_o$, the object's orientation.

The estimator consists of two subsystems that are running continuously in parallel: the wrench constraint and contact mode/geometry estimator (\secref{subsec:wrench_cone_est}), and the kinematic estimator (\secref{subsec:kinematic_est}). 
The former is not procedural: the generalized friction cone and contact mode/geometry can be estimated regardless of initial sequence of motions.
Kinematic estimation, on the other hand, requires a warm-start, in which the ground contact location is estimated first by enforcing sticking at that contact. This then allows the estimation of hand sliding $s_h$ and ground sliding $s_g$.


\subsection{Wrench constraint and contact mode/geometry}
\label{subsec:wrench_cone_est}
The complementarity conditions in equations
 \eqref{eq:slide_hand_1}-\eqref{eq:pivot_hand_2} and \eqref{eq:slide_gnd_3}-\eqref{eq:slide_gnd_4} allow us to estimate/regulate the contact mode and geometry by measuring/regulating the hand contact wrench. The first step is to estimate the wrench space boundaries on the left-hand side (LHS) of the complementarity (\eqref{eq:slide_hand_1}-\eqref{eq:pivot_hand_2} and \eqref{eq:slide_gnd_3}-\eqref{eq:slide_gnd_4}).
 
 \begin{figure}
    \centering
    \includegraphics[width=0.9\columnwidth]{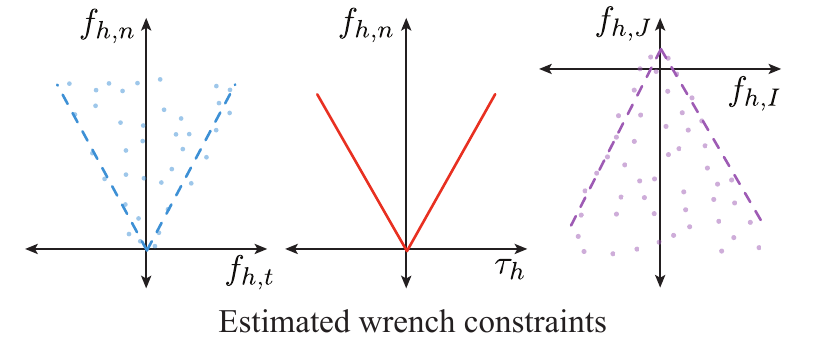}
    \vspace{-0.4cm}
    \caption{An illustration of estimated wrench cone constraints (i.e., LHS of equations  \eqref{eq:slide_hand_1}-\eqref{eq:pivot_hand_2} and \eqref{eq:slide_gnd_3}-\eqref{eq:slide_gnd_4}). Measured wrenches at the hand/ground are shown as blue/purple dots. A conservative estimate of the hand/ground friction constraints are shown in dashed blue/purple, and the prescribed torque constraint is shown in solid red.}
    \label{fig:est_example}
    \vspace{-0.6cm}
\end{figure}

We want to construct a conservative online estimate of the wrench boundaries that is robust to sensor noise and possible outliers.
A conservative estimate is desirable, as misclassifying a sliding/pivoting contact as a sticking/flush contact compromises both kinematic estimation and the controller's ability to enforce sticking/flush contact.  
Our estimator relies on the following: (a) that every measured wrench satisfies the wrench constraints, (b) the wrench boundaries are convex, (c) a conservative estimate is sufficient, and (d) the wrench constraints are constant in either the contact (LHS of \eqref{eq:slide_hand_1}-\eqref{eq:pivot_hand_2}) or world (LHS of \eqref{eq:slide_gnd_3}-\eqref{eq:slide_gnd_4}) frames.

\myparagraph{Hand contact friction constraints}
We rewrite the friction constraints at the hand contact \eqref{eq:slide_hand_1}-\eqref{eq:slide_hand_2} as:
\begin{align}
    \label{eq:hand_friction_cone}
    \mu_{meas}=|(f_{h,t}/(f_{h,n}+\epsilon))|< |f_{h,t}/(f_{h,n})| \le \mu_h,
\end{align}
where $\mu_{meas}$ is a lower-bound on $\mu_h$ and $\epsilon > 0$ ensures that $\mu_{meas}$ is both conservative and well-defined, even when the hand wrench is close to zero. To estimate $\mu_h$, we compute $\mu_{meas}$ for each measured wrench. 
We use all measured data to estimate a single $\mu_{meas}$ as the hand contact friction cone is symmetric.
We then use the LiveStats \cite{livestats} online quantile estimator to synthesize the time-history of wrench measurements into an estimate of the 99\% quantile of $\mu_{meas}$, which is our estimate for $\mu_h$. The quantile threshold ise used to modulate outlier rejection. We found that the 99\% quantile provides sufficient robustness. A schematic of the estimated hand friction constraints are shown in \figref{fig:est_example} (left). 

\myparagraph{Hand contact torque constraints}
We assume that the object surface always contains the length of the hand.
Consequently, the torque boundaries at the hand contact (LHS of \eqref{eq:pivot_hand_1}-\eqref{eq:pivot_hand_2}) are known apriori (\figref{fig:est_example}, center).
In the future, we intend to estimate the left and right torque boundaries during overhang or point contact between the hand and object. 


\myparagraph{Ground contact friction constraints} Unlike the hand contact friction constraints, the ground friction boundaries (LHS of \eqref{eq:slide_gnd_3}-\eqref{eq:slide_gnd_4}) cannot be rewritten in a highly structured form, as they depend on the unknown mass of the object. Instead, we directly estimate a convex hull of the force measurements $(f_{h,J},f_{h,I})$ as a conservative approximation of the ground contact friction constraints (\figref{fig:est_example}, right). This estimate is also robust to different ground angles. We compute an online estimate of the convex hull of these forces using the time-history of wrench measurements. This uses the same LiveStats \cite{livestats} online quantile estimator to infer a set of supporting hyperplanes of the convex hull. The intersections of these hyperplanes are used to identify candidate corners of the convex hull, which are then refined into the final approximation of the ground contact friction cone. A detailed explanation can be found in the appendix on our project page. 

\myparagraph{Contact mode and geometry estimation} Based on the complementarity conditions, we use our estimates of the wrench constraints to infer the contact mode of the system. 
We predict that the hand is sliding/pivoting if the measured wrench is in (near) violation of our estimates of the LHS of \eqref{eq:slide_hand_1}-\eqref{eq:slide_hand_2}/\eqref{eq:pivot_hand_1}-\eqref{eq:pivot_hand_2}, and in sticking/line contact otherwise. 
Similarly, we predict that object is sliding along the ground if the measured wrench is in (near) violation of our estimates of the LHS of \eqref{eq:slide_gnd_3}-\eqref{eq:slide_gnd_4} and is sticking otherwise. 
As discussed above, this estimation scheme is conservative: sliding/pivoting contact is rarely misclassified as sticking/flush contact, but the reverse misclassification can be frequent. 
This is intentional as undetected sliding/pivoting introduces error into the kinematic estimator and can result in the system moving into an unrecoverable state.



\subsection{Kinematics}
\label{subsec:kinematic_est}

The kinematic estimator synthesizes measurements from the time-history of robot proprioception and the current contact mode estimate into estimates of the object's orientation, as well as the hand and ground contact locations. 
This is done using a least-squares approach, where the update rule is selected based on the current contact mode estimate.
We exploit the kinematic constraints that sticking/line contacts place on the system's motion. These are the right-hand side (RHS) of the complementarity in \eqref{eq:slide_hand_1}-\eqref{eq:pivot_hand_2} and \eqref{eq:slide_gnd_3}-\eqref{eq:slide_gnd_4} .
%





When the hand and object are in line contact, their contact faces are parallel (i.e., $\Delta \theta = 0$, meaning $\theta_h = \theta_o$). In this case, we relate the hand pose to the ground contact as follows:
\begin{align}
\label{eq:kinematics}
-s_h \*{e_{h,t}} - d \*{e_{h,n}} + \*{r_o} = \*{r_h}, 
\end{align}
where $\*{r_h} = [x_h, y_h]^T$, $\*{r_o} = [x_o, y_o]^T$, and $d$ is the constant distance between the ground contact and the object face in contact with the hand (see \figref{fig:primitives}a). Due to line-contact, $\*{e_{h,n}}$ and $\*{e_{h,t}}$ are unit vectors normal and tangent to both hand and object contact surfaces. 
The location of the ground contact, $\*{r_{o}}$, is constant across periods of sticking point contact with the ground. During these periods, we regress $d$, $x_{o}$, and $y_{o}$ from the dot product of \eqref{eq:kinematics} with $\*{e_{h,n}}$:
\begin{align}
\label{eq:regression}
d (-1) + x_{o} (\*{I} \cdot \*{e_{h,n}}) + y_{o} (\*{J}\cdot \*{e_{h,n}}) = \*{r_h} \cdot \*{e_{h,n}}, 
\end{align}
where $\*{e_{h,n}}$ and $\*{r_h}$ are measured. Note that $s_g = x_o$. 

Once the ground contact location is inferred, we can estimate the sliding position of the hand $s_h$ from the hand's position during line-contact via the relation: 
\begin{align}
\label{eq:s_h_estimation}
s_h =  (\*{r_o}-\*{r_h}) \cdot \*{e_{h,t}}
\end{align}
During periods of sticking line contact at the hand (i.e., $\Delta \theta = \dot{s}_h = 0$), the hand and object move as a single rigid body. We solve \eqref{eq:kinematics} for $\mathbf{r_o}$ and use that to update $x_{o}, y_{o}$ during periods of sliding ground contact. Here we use the current hand pose ($\mathbf{r_h}$,  $\*{e_{h,n}}$, and $\*{e_{h,t}}$) as well as the most recent estimates of $d$ and $s_h$ before sliding at the ground contact is detected. Finally, when the object is in point contact with the hand and sticking with the ground, we can still use the system's kinematics to estimate $\theta_o$ and $s_h$. In all other contact mode combinations, the motion of the system is indeterminate, and we reset the estimator. 



\section{Object-agnostic control}
\label{sec:control}

\begin{figure}
    \centering
    \includegraphics[width=0.9\columnwidth]{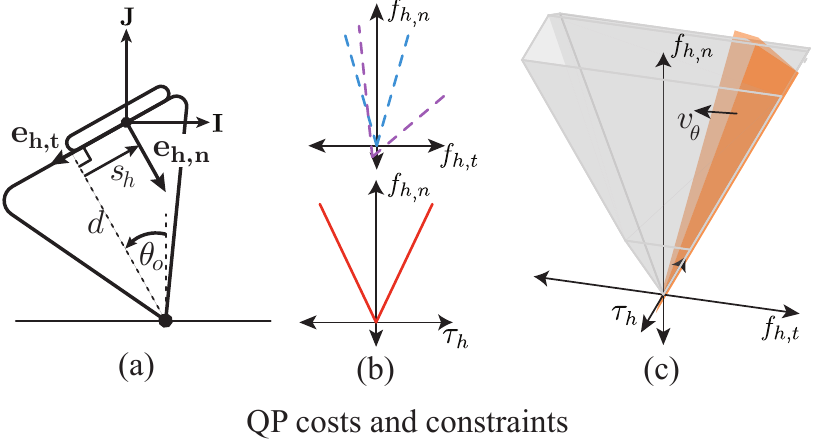}
    \vspace{-0.35cm}
    \caption{The QP cost and constraints. (a) A diagram showing the system state.  (b) The corresponding QP constraints for maintaining sticking at both contacts (top) and maintaining line-contact with the hand (bottom). The feasible region of the QP is the intersection of the estimated ground friction cone (dashed purple, top), hand friction cone (dashed blue, top), and prescribed hand torque cone (solid red, bottom). Since these constraints are drawn in the hand frame, the ground friction cone is rotated clockwise by $\theta_o$. (c) The boundary of the QP's 3D feasible region is shown in gray, the target direction associated with regulating the object's orientation, $\mathbf{v_{\theta}}$, is shown in black, and the intersection of the minimum-cost hyperplane defined by $\mathbf{v_{\theta}}$ and $\Delta \theta_o$ with the feasible region is shown in orange.}
    \label{fig:control_est_example}
    \vspace{-0.6cm}
\end{figure}

\begin{figure*}
    \centering
    \includegraphics[width=\textwidth]{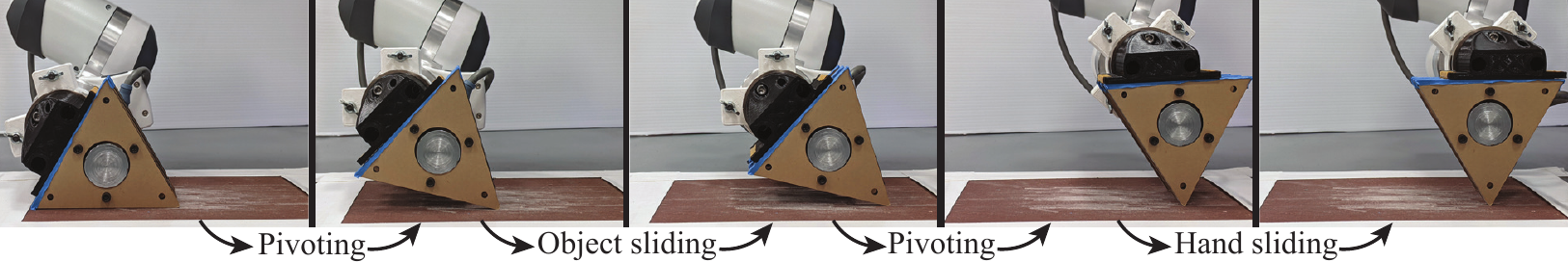}
    \vspace{-0.5cm}
    \caption{A contact configuration sequence during manipulation of an equilateral triangle with an addition of \SI{500}{\gram} mass on a high friction surface.}
    \label{fig:example_experiment}
    \vspace{-0.4cm}
\end{figure*}

The controller drives the system along admissible motion directions (a subset of $\Delta \theta_o$, $\Delta s_h$, $\Delta s_g$) and regulates the hand contact wrench to maintain the desired contact mode/geometry. 
The regulation of both forces and motions is achieved through a low-level impedance control layer.
We prefer an IFC over explicitly defining force and motion control subspaces as it is more tolerant of uncertainty in the kinematic estimates. 

For a quasi-static system, the low-level impedance controller can be approximated by the following compliance law: 
\begin{align}
\label{eq:impedance_law}
 \mathbf{w_h} =  \mathbf{K} (\mathbf{x_\text{tar}} -  \mathbf{x_h})
\end{align}
Here $\mathbf{x}_\text{tar}$ is the planar impedance target for the hand, $\mathbf{x_h}$ is the planar pose of the hand, $\*{K}$ is a stiffness matrix, and $\mathbf{w_h}$ is the wrench exerted by the robot. Our controller regulates both $\mathbf{x_h}$ and $\mathbf{w_h}$ by updating $\mathbf{x}_\text{tar}$ while keeping $\*{K}$ constant. 

The controller relies on the intuition that, for a fixed contact mode/geometry, moving the impedance target will only change the equilibrium state of the system along the admissible motion directions.
For the contact mode/geometry shown in \figref{fig:control_est_example}a, this is object rotation ($\Delta \theta_o$).
The controller can reduce object orientation error by moving the impedance target along the direction $\mathbf{v_{\theta}} = [-d, s_h, 1]^T $ in the hand contact frame. 
The product  $\Delta \theta_o \mathbf{v_{\theta}}$ corresponds to the predicted change in the hand pose for an incremental rotation by $\Delta \theta_o$ of the object about the estimated ground contact. 
The controller can also reduce error in $s_h$ and $s_g$ using similar reasoning, with $\mathbf{v_{h}} = [-1, 0, 0]^T$ in the hand frame and $\mathbf{v_{g}} = [1, 0, 0]^T$ in the world frame. Here $\Delta s_h \mathbf{v_{h}}$ ($\Delta s_g \mathbf{v_{g}}$) corresponds to incremental changes in the hand pose due to sliding by $\Delta s_h$ ($\Delta s_g$).

%

Regulation of the contact wrench ensures that the contact mode/geometry is maintained and the equilibrium state of the system changes as expected.
Maintaining the contact mode/geometry shown in \figref{fig:control_est_example}a, requires enforcing a strict inequality in the estimated LHS of \eqref{eq:slide_hand_1}-\eqref{eq:slide_hand_2} (i.e., sticking at the hand contact), \eqref{eq:pivot_hand_1}-\eqref{eq:pivot_hand_2} (i.e., flush hand contact),  and \eqref{eq:slide_gnd_3}-\eqref{eq:slide_gnd_4} (i.e., sticking at the ground contact). Planar projections of the estimated constraints are shown in \figref{fig:control_est_example}b.
%
As such, the specific wrenches acting on the object do not matter to the controller, so long as they satisfy the wrench constraints defined by the  contact mode/geometry. 



For this example, the controller computes the incremental change of the hand impedance target ($\Delta \mathbf{x_\text{tar}}$) as the solution to the following quadratic program (QP), visualized in \figref{fig:control_est_example}c: 
\begin{align}
\label{eq:example_qp_cost}
& \min_{\Delta \mathbf{w_h}, \Delta \mathbf{x_\text{tar}}} \  \big(\Delta \mathbf{x}_\text{tar}^T \*{v_{\theta}} - \Delta \theta_o)^2 \\
\label{eq:example_qp_iq_const}
 & \ \ \ s.t. \quad \ \  \mathbf{n}_j^T (\mathbf{w}_{h,meas} +  \gamma_j \Delta  \mathbf{w_h}) \leq b_j \quad \forall j \\
 \label{eq:example_qp_eq_const}
& \qquad \qquad \Delta \mathbf{w_h} =  \*{K} \Delta \mathbf{x_\text{tar}}
 \end{align}
The sum $\mathbf{w}_\text{meas} + \Delta  \mathbf{w_h}$ is the predicted wrench that the hand will exert after the impedance target has been incremented by $\Delta \mathbf{x_\text{tar}}$. The set of constraints \eqref{eq:example_qp_iq_const} correspond to the estimated value of the wrench space constraints discussed above plus a maximum on the normal force applied by the hand. The scaling factors $\gamma_j$ are used to amplify or attenuate wrench corrections. The equality constraint \eqref{eq:example_qp_eq_const} is an incremental approximation of \eqref{eq:impedance_law} that relies on the assumption that $\Delta \*{x_h}$ is small along the directions where the hand applies a wrench $\Delta \*{w_h}$ to the object (i.e., the constrained motion directions). Finally, as discussed, the cost \eqref{eq:example_qp_cost} reduces control error along the admissible motion directions by moving the impedance target along corresponding hand frame target directions.


 
The QP shown above can be applied to any of the contact modes/geometries discussed by modifying the cost \eqref{eq:example_qp_cost} and wrench constraints \eqref{eq:example_qp_iq_const}. The cost is generalized as:
\begin{align}
\label{eq:qp_cost_gen}
\alpha_0  || \Delta \mathbf{x_{tar}} ||^2 + \sum \alpha_i \big(\Delta \mathbf{x_{tar}}^T \mathbf{v}_i - \beta_i \Delta \epsilon_i \big)^2
\end{align}
Here $\Delta \epsilon_i$ are the errors along the admissible motion directions, the $|| \Delta \mathbf{x_{tar}} ||^2$ term regularizes the cost, and $\alpha_i$ and $\beta_i$ are controller gains. The constraints  \eqref{eq:example_qp_iq_const} are generalized by removing the constraint the prevents the desired motion. For example, we would remove the wrench constraint on the LHS of \eqref{eq:slide_hand_1} to allow for positive sliding at the hand contact. The incremental change of the hand impedance target is then computed by minimizing this cost and the appropriately selected wrench constraints subject to \eqref{eq:example_qp_eq_const}.  
\vspace{-0.2cm}

\section{Joint estimation and control}
\label{sec:joint_est_and_cont}


The estimator and controller work together to manipulate an unknown object. The hand starts in line-contact with the object, and consequently, its orientation is known. We initialize the estimator with conservative guesses of the friction constraints (i.e., small friction coefficients). The object geometry, relative pose of the hand, and pivot location are unknown. We manipulate the object in a reactive fashion through a hand-scripted sequence of contact configurations. 

Initially, we focus on exploration to improve our estimates of the system's kinematics and wrench constraints. We use the controller described above with a naive guess of $\mathbf{v_{\theta}} = [-d_{guess}, 0, 1]^T$, and noting that $\*{v_h}$ and $\*{v_g}$ are already geometry-agnostic. While initial commanded motions are limited by our conservative estimate of the friction constraints, we more accurately estimate each constraint by commanding sliding in that direction. For example, commanding right sliding at the hand (i.e, $\dot{s}_h^+ > 0$) removes the LHS of \eqref{eq:slide_hand_1} from the QP and drives the measured wrench towards the true constraint boundary. This allows the estimator to more accurately estimate that constraint. We also command object rotations to estimate the pivot location.

As the robot manipulates the object, its estimates of the wrench constraints and system kinematics become more accurate. This improves the controller's performance by refining $\*{v_\theta}$ and expanding the feasible region of the QP.

\section{Experiments and Results}

\begin{figure}[t]
    \centering
    \includegraphics[width=0.9\columnwidth]{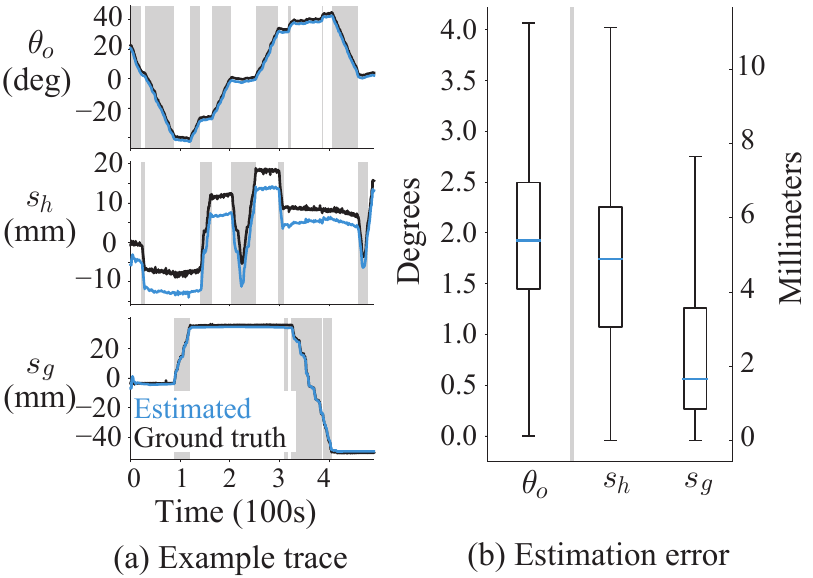}
    \vspace{-0.3cm}
    \caption{(a) Ground truth (black) versus estimated (blue) time traces for object orientation (top), hand sliding position (middle), and object sliding position (bottom) during manipulation of a triangle with no additional mass. Highlighted gray regions indicate when the corresponding variable was commanded to change. (b) Box and whisker plots show absolute estimation error for all ten trials. The median is in blue, top and bottom edges of the box are the 75\% and 25\% quantiles, and each whisker is 1.5 $\times$ the box length. Outliers constitute 5\% of the data and are not shown. }
    \vspace{-0.2cm}
    \label{fig:estimator_traces}
\end{figure}

\begin{figure}[t]
    \centering
    \includegraphics[width=0.9\columnwidth]{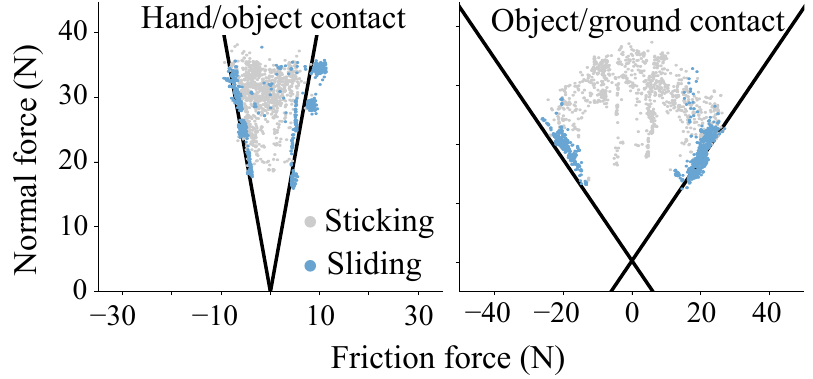}
    \vspace{-0.3cm}
    \caption{Estimated friction cone boundaries (black) for the hand (left) and ground (right) contacts during the same trial as in \figref{fig:estimator_traces}. We superimpose measured wrenches, which are colored gray for sticking and blue for sliding based on the contact mode measured via the ground truth. }
    \label{fig:wrench_cone_data}
    \vspace{-0.5cm}
\end{figure}

\label{sec:experiments_and_results}
\begin{figure}[t]
    \centering
    \includegraphics[width=0.9\columnwidth]{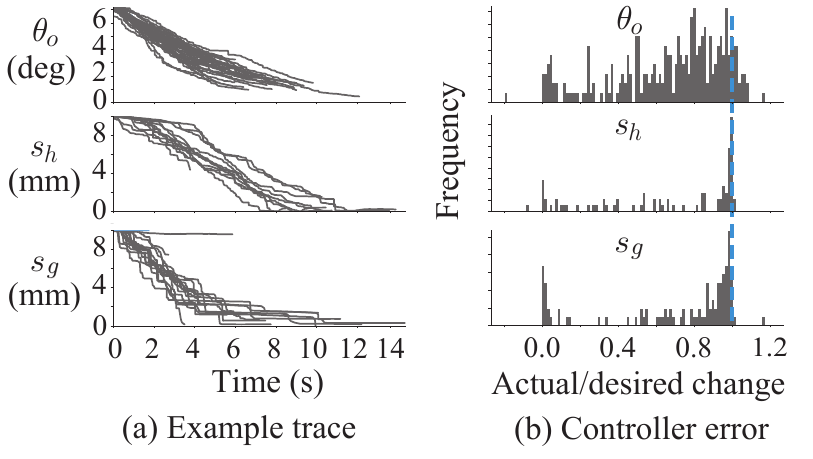}
     \vspace{-0.3cm}
    \caption{(a) Time traces of the control error during incremental motions along the admissible motions directions for the same trial as in \figref{fig:estimator_traces}. (b) A histogram of the actual change in pose normalized by the commanded change in pose along admissible motion directions across all ten trials. A value of one (dashed blue line) indicates that the commanded change has been fully realized.}
    \label{fig:controller_conv}
    \vspace{-0.1cm}
\end{figure}

We conduct several experiments to verify our framework using Franka Emika's Panda robot in impedance control mode. The estimators and controller run at \SI{100}{\hertz} and \SI{30}{\hertz}. The controller parameters, i.e., $\alpha$, $\beta$, and $\gamma$, are manually tuned and then fixed for all trials.

\myparagraph{Object-agnostic manipulation} We test our framework's ability to manipulate a set of unknown polygons and average the results across ten experiments. For each experiment, the robot manipulates either an equilateral triangle, a square, a rectangle, or regular hexagon or pentagon. The objects' side-length and mass vary between \SIrange[]{7}{19}{\centi\meter} and \SIrange[]{200}{450}{\gram}. For two experiments, we add a \SI{500}{\gram} mass to the triangle and pentagon. The coefficient of friction between the hand/object and object/ground is measured as approximately 0.5 and 0.8, respectively. The video shows additional experiments using friction surfaces with lower coefficients of friction and non-polygonal objects. The estimator and controller are initialized as described in \secref{sec:joint_est_and_cont}, and the manipulator moves through a long sequence of contact configurations during each trial.  We command changes in $\theta_o$, $s_g$ and $s_h$ in increments of \SI{7.2}{\degree}, \SI{10}{\milli\meter}, and \SI{10}{\milli\meter}, respectively. A portion of this sequence is shown in \figref{fig:example_experiment}. The most commonly observed failure mode is unintentional slipping at the external contact.

We compare the performance of our kinematic estimator against ground truth provided using the AprilTag vision system \cite{olson2011apriltag}. We show an example time trace in \figref{fig:estimator_traces}a, and present the estimation error statistics for all ten trials in  \figref{fig:estimator_traces}b. These results indicate that kinematic estimation is sufficiently accurate for control purposes. We also compare the wrench cone constraint estimates to the contact mode as detected via the ground truth in \figref{fig:wrench_cone_data}. As expected, the measured wrench is near the edge of the friction cone when sliding occurs. We finally analyze the ability of the controller to execute motions along the admissible directions $\Delta \theta_o$, $\Delta s_h$, and $\Delta s_g$ in \figref{fig:controller_conv}. We find that the controller drives the system towards the commanded state.

\begin{figure}[t]
    \centering
    \includegraphics[width=0.9\columnwidth]{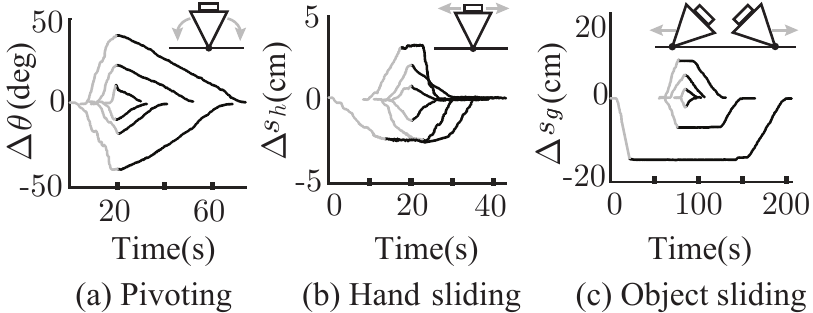}
    \vspace{-0.3cm}
    \caption{A demonstration of pose regulation along the admissible motion directions. Time traces of the ground truth error in the state variable being actively regulated are shown. Gray lines correspond to a manually applied perturbation, and black lines show the controller's response.}
    \label{fig:perturbation}
    \vspace{-0.6cm}
\end{figure}

\myparagraph{Perturbation rejection} We also perform experiments to test the controller's ability to reject perturbations along the three admissible motion directions (\figref{fig:perturbation}). We find that the system is able to recover from large perturbations in $\theta_o$, $s_h$, and $s_g$ as long as the initial state can be reached from the perturbed state under the given contact geometry and mode. For instance, for the given system parameters and initial conditions, ground sliding regulation (bottom, \figref{fig:perturbation}) can only correct perturbations in one direction.

\section{Conclusion}
\label{sec:conclusion}

We demonstrate that objects can be manipulated by estimating and controlling their contact configurations. This allows for object manipulation without knowledge of its pose, inertia, or shape, though we note that estimating inertia is more important for dynamic manipulation. We use this to manipulate unknown planar objects, and also posit that contact configuration regulation can be used more generally. 

One limitation of this approach is our inability to estimate the length and location of patch contacts between the object and the hand or ground. The key difficulty is that, unlike the friction parameters, the length and location of these patches will change based on the motion of the hand and object. We plan to address this in the future by incorporating kinematic feedback into the wrench cone estimator. Another limitation is the use of a simple line-contact hand limits control authority. A more articulated hand (e.g., a parallel-jaw gripper) can overcome this limitation, and we plan to extend our framework to apply to hand/object contact geometries induced by different hands.

Finally, we also plan to develop a higher level planning framework that automatically sequences contact configurations and intermediate target states to (a) move between states that require sequencing multiple contact configurations, and (b) to more efficiently explore the state space and estimate the kinematic parameters and wrench constraints. 


\bibliographystyle{IEEEtran}
\bibliography{references}

\end{document}